\documentclass[10pt,english]{article}
\usepackage{graphicx}
\usepackage[T1]{fontenc}
\usepackage[latin9]{inputenc}
\usepackage[margin=1.0in]{geometry}
\usepackage{float}
\usepackage{amsmath}
\usepackage{amsbsy}
\usepackage{setspace}
\usepackage{amssymb}
\usepackage{esint}
\usepackage{cite}
\newfloat{algorithm}{tbp}{loa}
\floatname{algorithm}{Algorithm}
\usepackage{algorithmic}

\newlength\myindent
\makeatletter

\floatstyle{ruled}
\newfloat{algorithm}{tbp}{loa}
\floatname{algorithm}{Algorithm}

\usepackage{babel}

\begin{document}

\title{Randomized Kernel Methods for \\ Least-Squares Support Vector Machines}

\author{M. Andrecut}

\date{June 23, 2016}

\maketitle
{

\centering Calgary, Alberta, T3G 5Y8, Canada

\centering mircea.andrecut@gmail.com

} 
\bigskip 
\begin{abstract}
The least-squares support vector machine is a frequently used kernel method for non-linear regression and classification tasks. 
Here we discuss several approximation algorithms for the least-squares support 
vector machine classifier. The proposed methods are based on randomized block kernel 
matrices, and we show that they provide good accuracy and reliable scaling for multi-class 
classification problems with relatively large data sets. Also, we present several numerical 
experiments that illustrate the practical applicability of the proposed methods.


\end{abstract}
\bigskip 

\section{Introduction}

Kernel methods play an important role in solving various machine learning problems, 
such as non-linear regression and classification tasks \cite{key-1}. 
The kernel methods rely on the so called "kernel trick", which transforms a given 
non-linear problem into a linear one by using a kernel function $\Omega(x,x')$, which 
is a similarity function defined over pairs of input data points $(x,x')$. This way, 
the input data $x$ is mapped into a high dimensional (or even infinite-dimensional) 
feature space $\phi(x)$, where the inner product $\langle\cdot,\cdot\rangle$ can be calculated with a 
positive definite kernel function (that is satisfying Mercer's condition)\cite{key-2}: 
\begin{equation}
\Omega(x,x') = \langle \phi(x),\phi(x')\rangle.
\end{equation}
Therefore, the mapping into the high dimensional feature space is done implicitly, 
without the need to explicitly map the data points $\phi(x)$. Also, assuming that 
$\chi = \{x_n|n=1,\dots,N \}$ is the training data, then using the 
Representer Theorem any non-linear function $f:\chi \rightarrow \mathbb{R}$ can 
be expressed as a linear combination of kernel products evaluated on the 
training data points \cite{key-1}:
\begin{equation}
f(x) = \sum_{n=1}^{N} a_n \Omega(x,x_n).
\end{equation}

The main methods for the kernel classification problems are the support vector machines 
(SVM)\cite{key-3} and the least-squares support vector machines (LS-SVM)\cite{key-4}. 
In this paper we focus on the LS-SVM classifier, where the main difficulty is the $O(N^3)$ training complexity, 
where $N$ is the size of the training data set. Because of this high complexity, the 
LS-SVM method is not a suitable candidate for applications with large data sets.
Here, we discuss several approximation methods using randomized block kernel matrices, 
that significantly reduce the complexity of the problem. 
The proposed methods are based on the Nystr\"{o}m\cite{key-5}, 
Kaczmarz\cite{key-6} and Matching Pursuit\cite{key-7} algorithms, and we show that they 
provide good accuracy and reliable scaling to relatively large multi-class classification 
problems. 

\section{Kernel LS-SVM classifier}

In the binary classification setting the kernel SVM method constructs an optimal separating 
hyperplane (with the maximal margin) between the two classes in the feature space $\phi(x)$.  
The training problem is represented as a (convex) quadratic programming problem involving 
inequality constraints, which has a global and unique solution \cite{key-3}.

The kernel LS-SVM method simplifies the optimization problem by considering 
equality constraints only, such that the solution is obtained by solving a system of linear 
equations \cite{key-4}. With these modifications, the problem is equivalent to a ridge 
regression formulation with binary targets $\pm 1$. Also, the kernel LS-SVM allows us to treat 
the multi-class classification task in a much simple and compact way. 
More exactly, we assume that $K$ classes are encoded using the standard 
basis in the $\mathbb{R}^K$ space. Therefore, if $x_i \in C_k$ is an input in class $C_k$, 
then the output $y_i$ is encoded by a binary row vector with 1 in the $k$th position and 0 
in all other positions:
\begin{equation}
x_i \in C_k \; \Leftrightarrow \; y_{ij} = 
\begin{cases}
		1 & \text{if }\; j=k \\
		0 & \text{otherwise}
\end{cases}.	
\end{equation}

Thus, for the input data 
$\{(x_i,y_i)\vert x_i \in \mathbb{R}^M, y_i \in \mathbb{R}^K, i=1,\dots,N\}$  
and the feature mapping function $\phi(x)$ we consider the following optimization problem:\cite{key-4}
\begin{equation}
\min_{w,b,\varepsilon} S(w,b,\varepsilon) = 
\frac{1}{2} \sum_{j=1}^K \left(  \Vert w^{(j)} \Vert^2 + \gamma \sum_{i=1}^N \varepsilon_{ij}^2 \right) 
\end{equation}
such that:
\begin{equation}
\langle \phi(x_i), w^{(j)} \rangle + b_j + \varepsilon_{ij} = y_{ij}, \quad i=1,\dots,N;\quad j=1,\dots,K, 
\end{equation}
where $b_j$ is the bias coefficient, $w^{(j)}$ is the 
vector of weights corresponding to the class $j$, and $\varepsilon_{ij}$ is the approximation error. 
The objective function $S$ is a sum of the least-squares error and a regularization term, and therefore it 
corresponds to a multi-dimensional ridge regression problem with the regularization parameter $\gamma$. 

In the primal weight space the multi-class classifier takes the form:
\begin{equation}
x \in C_k \; \Leftrightarrow \; k = \text{arg} \max_{j=1,\dots,K} g_j(x)
\end{equation}
where $g_J(x)$ is the nonlinear softmax function:

\begin{equation}
g_j(x) = \frac{\exp \left(  \langle \phi(x), w^{(j)} \rangle + b_j \right) }
{\sum_{i=1}^K \exp \left(  \langle \phi(x),  w^{(i)} \rangle + b_i \right) }
\end{equation}

Because $\phi(x)$ may be infinite dimensional we seek a solution in the dual space $\phi(x)$ by introducing the Lagrangian:
\begin{equation}
L(w,b,\varepsilon,a) = S(w,b,\varepsilon) - \sum_{i=1}^N \sum_{j=1}^K a_{ij} \left[ \langle \phi(x_i), w^{(j)} \rangle + b_j + \varepsilon_{ij} - y_{ij} \right], 
\end{equation}
with the Lagrange multipliers $a_{ij} \in \mathbb{R}$. 

The optimality conditions for the minimization problem are:
\begin{align}
\frac{\partial L}{\partial w^{(j)}} = 0 \quad &\Rightarrow \quad w^{(j)} = \sum_{n=1}^N a_{nj} \phi(x_n) \\
\frac{\partial L}{\partial b_j} = 0 \quad &\Rightarrow \quad \sum_{i=1}^N a_{ij} = 0 \\ 
\frac{\partial L}{\partial \varepsilon_{ij}} = 0 \quad &\Rightarrow \quad a_{ij} = \gamma \varepsilon_{ij} \\ 
\frac{\partial L}{\partial a_{ij}} = 0  \quad &\Rightarrow \quad \langle \phi(x_i), w^{(j)} \rangle + b_j + \varepsilon_{ij} - y_{ij} = 0
\end{align}
By eliminating $w^{(j)}$ and $\varepsilon_{ij}$ we obtain:
\begin{equation}
\sum_{n=1}^N \left[ \Omega(x_i, x_n) + \gamma^{-1} \delta_{in} \right] a_{nj} + b_j = y_{ij}, \quad i=1,\dots,N,\; j=1,\dots,K,
\end{equation}
where we applied the condition $\Omega(x_i, x_n) = \langle \phi(x_i), \phi(x_n) \rangle$, 
and $\delta_{in}$ is Kronecker's delta: $\delta_{in}=1$ if $i=n$, 
and $\delta_{in}=0$ otherwise. 

Therefore, in the dual space the multi-class classifier takes the form:
\begin{align}
x \in C_k \; \Leftrightarrow \; k = \text{arg} \max_{j=1,\dots,K} g_j(x), \\
g_j(x) = \frac{\sum_{n=1}^N \exp \left( \Omega(x, x_n) a_{nj} + b_j \right)}
{\sum_{i=1}^K \sum_{n=1}^N \exp \left(\Omega(x, x_n) a_{ni} + b_i \right)}.
\end{align}

One can see that the above system of equations (13) is equivalent to $K$ independent systems of equations with 
binary targets $y_{ij} \in \{0,1\}$. Each system can be written in the following equivalent form:
\begin{equation}
\begin{bmatrix}
    0 & \; & u^T  \\
    u & \; &\Omega + \gamma^{-1}I 
\end{bmatrix}
\begin{bmatrix}
    b_j \\
    a^{(j)} 
\end{bmatrix} = 
\begin{bmatrix}
    0 \\
    y^{(j)} 
\end{bmatrix}, \quad j=1,\dots,K,
\end{equation}
where $I$ is the $N\times N$ identity matrix, $u = [1,\dots,1]^T$ is an $N$ 
dimensional vector with all the components equal to 1, 
$a^{(j)}=[a_{1j},\dots,a_{Nj}]^T$ and $y^{(j)}=[y_{1j},\dots,y_{Nj}]^T$ are 
the weight and respectively the target column vectors for the class $j=1,\dots,K$.
Each system has $N+1$ linear equations with $N+1$ unknowns, 
and requires the inversion of the same matrix:
\begin{equation}
\Theta = 
\begin{bmatrix}
    0 & \; & u^T  \\
    u & \; &\Omega + \gamma^{-1}I 
\end{bmatrix}.
\end{equation}
The above $K$ systems can be written in a compact matrix form as following:
\begin{equation}
\Theta W = Z
\end{equation}
where $W$ and $Z$ are $(N+1) \times K$ matrices with the columns:
\begin{equation}
w^{(j)}(t) = 
\begin{bmatrix}
    b_j(t) \\
    a^{(j)}(t) 
\end{bmatrix}, \;
z^{(j)} = 
\begin{bmatrix}
    0 \\
    y^{(j)} 
\end{bmatrix}, \quad j=1,\dots,K.
\end{equation}
Despite these nice mathematical properties, the complexity of the problem is $\sim O(N^3)$, 
and it is implied by the matrix inversion. 

\section{Randomized kernel methods}

Our goal is to reduce the complexity of the kernel LS-SVM classification method by using iterative approximations based on 
randomized block kernel matrices. More exactly we use the well known Nystr\"{o}m  
method,\cite{key-5} and we introduce new methods based on the  
Kaczmarz\cite{key-6} and the Matching Pursuit\cite{key-7} algorithms.  

\subsection{Randomized Nystr\"{o}m method}

The Nystr\"{o}m method is probably the most popular low rank approximation of the matrix $\Theta$.\cite{key-5,key-8}  
Since the kernel matrix is a positive definite symmetric matrix we can find its eigenvalue decomposition:
\begin{equation}
\Theta = V\Lambda V^T
\end{equation}
where $\Lambda$ is the diagonal matrix of the eigenvalues, and $V$ is the corresponding matrix of the eigenvectors.
Using the Sherman-Morrison formula\cite{key-9} one can show that the solution of the linear system of equations can be written as:
\begin{equation}
w^{(j)} = \gamma \left[ z^{(j)} - V \left(\gamma^{-1}I + \Lambda V^T V \right)^{-1} \Lambda V^T z^{(j)} \right] 
\end{equation}
To reduce the computation one can consider only a small selection $\Theta_{(J,J)} \in \mathbb{R}^{J \times J}$, $J \ll N$, 
based on a random sample of the training data. The eigenvalue decomposition of $\Omega_{(J,J)}$ is therefore:
\begin{equation}
\Theta_{(J,J)} = \bar{V} \bar{\Lambda} \bar{V}^T
\end{equation}
One can show that the following relations exist between the eigenvalues and eigenvectors of the two matrices 
$\Theta$ and respectively $\Theta_{(J,J)}$:
\begin{align}
\lambda_n &= \frac{N}{J} \bar{\lambda}_n \\
u_n &= \sqrt{\frac{N}{J}} \frac{1}{\bar{\lambda}_n} \Theta_{(J,J)} \bar{u}^{(n)}, \quad n = 1,\dots,N.
\end{align}
such that:
\begin{equation}
\Theta \simeq \Theta_{(N,J)} \Omega_{(J,J)}^{-1} \Theta_{(J,N)}
\end{equation}
where $\Theta_{(N,J)}$ and  $\Theta_{(J,N)}$ are the $N \times J$ and respectively 
the $J \times N$ block submatrices taken from $\Theta$, corresponding 
to the randomly selected $J$ columns. 
Therefore, in this case only a matrix of size $J \times J$ needs to be inverted, which simplifies the 
computation. While this method seems to scale as $O(J^2(N+1))$ it has the difficulty of optimally choosing  
the $J$ support vectors from the training data set. It has been shown that an optimal method 
for choosing the $J$ support vectors is based on the maximization of the Renyi entropy of the the matrix 
$\Theta_{(J,J)}$.\cite{key-10} 
Therefore, one can use a heuristic greedy algorithm to search for the best set of $J$ support vectors, 
that maximize the Renyi entropy of $\Theta_{(J,J)}$, however this approach becomes difficult in the 
case of large data sets. 

Here we consider a much simpler approach, based on a committee machine made of $T$ classifiers. 
Let us assume that the classifiers are characterized by the following randomly selected matrices: 
$\Theta_{(J,J)}^{(t)}$, $\Theta_{(N,J)}^{(t)}$, $\Theta_{(J,N)}^{(t)}$, $t=1,\dots,T$.
The random sampling can be done without replacement, such that after $\lfloor (N+1)/J \rfloor$ 
selections all the training data is used. 
An efficient randomization strategy would be to use a random shuffling function which generates a random permutation 
$r(N) = \{r(1),\dots,r(N) \}$ of the set $\{1,\dots,N\}$,  
and then to select contiguous subsets with size $J$ from $r(N)$.  
After the algorithm consumes all the $\lfloor (N+1)/J \rfloor$ subsets one can re-shuffle $r(N)$ in order to create another 
randomized list. 

For each classifier we solve the system:
\begin{equation}
\Theta_{(N,J)}(t) \Theta_{(J,J)}^{-1}(t) \Theta_{(J,N)}(t) W(t) = Z. 
\end{equation}
The solution is given by:
\begin{equation}
W(t) = \Theta^{\dagger}_{(J,N)}(t) \Theta_{(J,J)}(t) \Theta^{\dagger}_{(N,J)}(t)Z, 
\end{equation}
where $\Theta^{\dagger}_{(N,J)}(t)$ and $\Theta^{\dagger}_{(J,N)}(t)$ are the Moore-Penrose pseudo-inverse matrices of the 
random block matrices $\Theta_{(N,J)}(t)$ and $\Theta_{(J,N)}(t)$, which 
can be calculated at a lower cost of $O(J^2N)$. 
In the end we aggregate the weights of the classifiers as following:
\begin{equation}
W = \sum_{t=1}^{T} W(t),
\end{equation}
and the classification process is performed using the equations (14) and (15).

\subsection{Randomized Kaczmarz method}

The Kaczmarz method\cite{key-6} is a popular solver for overdetermined linear systems of equations of the form:
\begin{equation}
Ax = b, \quad A \in \mathbb{R}^{N \times M}, \; x \in \mathbb{R}^{M}, \; b \in \mathbb{R}^{N}, 
\end{equation}
with $N\geq M$. The method has numerous applications in computer tomography and image reconstruction from projections. 
Assuming that $x(0)$ is an initial estimation of the solution, the algorithm proceeds cyclically and at each step it 
projects the estimate $x(t)$ onto a subspace normal to the row $a_{(i)}$ of $A$, such that:
\begin{equation}
x(t+1) = x(t) + \frac{b_i - \langle a_{(i)}, x(t) \rangle}{\Vert a_{(i)}\Vert^2} a_{(i)}, \quad i = t (\text{mod}N) + 1.
\end{equation}

Because of its inherent cyclical definition, the convergence of the method depends on the order of the rows.
A simple method to overcome this problem is to select the rows of $A$ in a random order:

\begin{equation}
x(t+1) = x(t) + \frac{b_{r(i)} - \langle a_{r(i)}, x(t) \rangle}{\Vert a_{r(i)}\Vert^2} a_{r(i)}, 
\end{equation}
where $r(i) \in \{1,\dots,N \}$ is a random function. It has been shown that if the rows are selected with a probability: 
\begin{equation}
p(i) = \Vert  a_{r(i)} \Vert^{2} \Vert A \Vert_F^{-2},\quad i=1,\dots,N,
\end{equation}
where $\Vert \cdot \Vert_F$ if the Frobenius norm, then the algorithm converges in expectation exponentially, with a rate 
independent on the number of equations:
\begin{equation}
\mathbb{E}(\Vert x(t) - x \Vert)^2 \leq (1 - \kappa(A)^{-2})^t \Vert x(0) - x \Vert,
\end{equation}
where
\begin{equation}
\kappa(A) = \sigma_{max}(A) \sigma_{min}^{-1}(A),
\end{equation}
is the condition number of $A$, with $\sigma_{max}(A)$ and $\sigma_{min}(A)$ the maximal and minimal singular values of $A$ respectively.\cite{key-11} 
This remarkable results shows that the estimate $x(t)$ converges exponentially fast to the solution, in just $O(N)$ iterations. Also, since 
each iteration requires $O(N)$ time steps, the overall complexity of the method is $O(N^2)$, which is much smaller than $O(MN^2)$ required 
by the standard approach. Another big advantage is the low memory requirement, since the method only needs the randomly chosen row 
at a given time, and it doesn't operate with the whole matrix $A$. This makes the method appealing for very large systems of equations. 

It is interesting to note that the matrix $A$ can be "standardized", such that each of its rows has unit Euclidean norm:
\begin{equation}
\hat{A}x = \hat{b},
\end{equation}
where
\begin{align}
\hat{a}_{(i)} &= \Vert a_{(i)} \Vert^{-1} a_{(i)} \\
\hat{b_i} &= \Vert a_{(i)} \Vert^{-1} b_i.
\end{align}
Therefore, by using this simple standardization method, the random row selection can be done with uniform probability, since all the rows have 
an identical norm $\Vert \hat{a}_{(i)} \Vert =1$, $i=1,\dots,N$.
In this case the iterative equation takes the form:
\begin{equation}
x(t+1) = x(t) + [\hat{b}_{r(i)} - \langle \hat{a}_{r(i)}, x(t) \rangle ] \hat{a}_{r(i)}, 
\end{equation}
and $r(i) \in \{1,\dots,N \}$ is a uniformly distributed random function.

In the case of our multi-class classification problem, we prefer a randomized block version of the Kaczmarz method, in order 
to use more efficiently the computing resources. Also we assume that the system of equations is standardized using the above described procedure. 
In this case at each iteration we randomly select a subset $s(t)$ of the row indexes of $A$, 
with a size $J=\vert s(t) \vert \ll N$, and we project the current estimate $x(t)$ onto a subspace normal to these rows:
\begin{equation}
x(t+1) = x(t) + \hat{A}_{s(t)}^\dagger [\hat{b}_{s(t)} - \hat{A}_{s(t)} x(t)],
\end{equation}
where:

\begin{equation}
\hat{A}_{s(t)}^\dagger = \hat{A}_{s(t)}^T \left[  \hat{A}_{s(t)}\hat{A}_{s(t)}^T \right] ^{-1}
\end{equation}
is the Moore-Penrose pseudo-inverse of the matrix $\hat{A}_{s(t)}$, and $\hat{b}_{s(t)}$ is the subvector of $\hat{b}$ 
with the components from $s(t)$. 

The above equation is equivalent to:
\begin{equation}
x(t+1) = x(t) + \hat{A}_{s(t)}^\dagger [\hat{A}_{s(t)}x - \hat{A}_{s(t)} x(t)],
\end{equation}
where $x$ is the exact solution of the system. Therefore we have:
\begin{equation}
x(t+1) = x(t) + \hat{A}_{s(t)}^\dagger \hat{A}_{s(t)} [x - x(t)].
\end{equation}
This can be rewritten as:
\begin{equation}
x(t+1) - x = x(t) - x - P^{\perp}_{s(t)} [x(t) - x].
\end{equation}
where $P^{\perp}_{s(t)} = \hat{A}_{s(t)}^\dagger \hat{A}_{s(t)}$.
Since $P^{\perp}_{s(t)}$ is an orthogonal projection we have:
\begin{equation}
\Vert x(t+1) - x \Vert^2 =  \Vert x(t) - x \Vert^2 - \Vert P^{\perp}_{s(t)} [x(t) - x] \Vert \leq  \Vert x(t) - x \Vert^2,
\end{equation}
and the algorithm is guaranteed to converge. In fact, for $J=\vert s(t) \vert = 1$, the randomized block algorithm is equivalent 
to the simple (one row at a time) randomized algorithm, which guarantees exponential convergence.\cite{key-12}

In our multi-class classification setting we solve simultaneously $K$ systems of equations, corresponding to the $K$ different classes: 
\begin{equation}
\Theta w^{(j)} = z^{(j)}, \quad j = 1,\dots,K.
\end{equation}
Therefore, the iterative equations are:
\begin{equation}
w^{(j)}(t+1) = w^{(j)}(t) + \hat{\Theta}_{s(t)}^\dagger  [\hat{z}^{(j)}_{s(t)} - \hat{\Theta}_{s(t)} w^{(j)}(t) ], \quad j=1,\dots,K,
\end{equation}
where $\hat{\Theta}_{s(t)}$ and $\hat{z}^{(j)}_{s(t)}$ are the standardized versions of the randomly selected block matrix $\Theta_{s(t)}$ 
and the corresponding target subvector $z^{(j)}_{s(t)}$. These equations can be written compactly in a matrix form as following:
\begin{equation}
W(t+1) = W(t) + \hat{\Theta}_{s(t)}^\dagger [\hat{Z}_{s(t)} - \hat{\Theta}_{s(t)} W(t) ].
\end{equation}
The iteration stops when no significant improvement for $W$ is made, or the number of iterations reach a maximum accepted value.
Once the matrix $W$ containing the weights $a_{ij}$ and the bias $b_{j}$ is calculated, one can use the multi-class classifier defined by the equations (14)-(15), 
in order to classify any new $x \in \mathbb{R}^M$ data sample.

\subsection{Randomized Matching Pursuit method}

The Matching Pursuit (MP) method is frequently used to decompose a given signal into a linear expansion of functions 
selected from a redundant dictionary of functions.\cite{key-7} 
Thus, given a signal $f \in \mathbb{R}^N$, we seek a linear  
expansion approximation:
\begin{equation}
f^* = \sum_{m=1}^M \alpha_{s(m)} \psi^{s(m)},
\end{equation}
where $\psi^{s(m)} \in \mathbb{R}^N$ are the columns of the redundant dictionary $\Psi \in N \times M$, with $M \geq N$. 
Here, $s(m) \in \{1,\dots,M\}$ is a selection function which returns the index of the column from the dictionary.
Thus, we solve the minimization problem:
\begin{equation}
\min_{\alpha} \Vert f - f^* \Vert^2.
\end{equation}
Starting with an initial approximation $f^*_0 = 0$, at each new step the algorithm iteratively selects a new column $\psi^{s(m)}$ 
from the dictionary, in order to reduce the future residual $r_{m} = f - f^*_{m}$. Therefore, from $f^*_{m-1}$ one can build a new approximation:
\begin{equation}
f_{m}^* = f^*_{m-1} + \alpha_{s(m)} \psi^{s(m)}
\end{equation}
by finding $\alpha_{s(m)}$ and $\psi^{s(m)}$ that minimizes the residual error:
\begin{equation}
\Vert r_{m} \Vert^2 = \Vert f - f^*_{m} \Vert^2 = \Vert r_{m-1} - \alpha_{s(m)}\psi^{s(m)} \Vert^2.
\end{equation}
The minimization condition:
\begin{equation}
\frac{\partial \Vert r_{m-1} - \alpha_{s(m)}\psi^{s(m)} \Vert^2}{\partial \alpha_{s(m)}} = 0
\end{equation}
gives:
\begin{equation}
\alpha_{s(m)} = \Vert \psi^{s(m)} \Vert^{-2} \langle \psi^{s(m)}, r_{m-1} \rangle,
\end{equation}
and therefore we have:
\begin{equation}
\Vert r_{m} \Vert^2 = \Vert r_{m-1} \Vert^2 - \left[  \Vert \psi^{s(m)} \Vert^{-1} \langle \psi^{s(m)}, r_{m-1} \rangle \right]^2 \leq \Vert r_{m-1} \Vert^2.
\end{equation}
The index of the best column that minimizes the residual is given by:
\begin{equation}
j = \text{arg} \max_{m=1,\dots,M} \left[  \Vert \psi^{s(m)} \Vert^{-1} \langle \psi^{s(m)}, r_{m-1} \rangle \right]^2. 
\end{equation}

The MP algorithm can be easily extended for kernel functions.\cite{key-12} 
However, for large data sets this approach is not quite efficient due to the increasing time required by the search for the best column. 
We notice that according to (54) the algorithm converges even if the selection of the column is randomly done.  
Therefore, here we prefer this weaker form, where the functions from the redundant dictionary are simply selected randomly.

In our case, we consider that the dictionary corresponds to the matrix $\Theta$, and as before we consider a block matrix formulation of the weak Matching Pursuit algorithm.
Initially the matrix $W$ is empty, $W(0)=0$, and the residual is set to $R(0) = Z$. At each iteration step 
we select randomly (without replacement) $J$ columns from $\Theta$, which we denote as the $N \times J$ matrix $\Theta_{s(t)}$, 
where $s(t)$ is the random list of columns. Thus, at each iteration step we solve the following system:
\begin{equation}
\Theta_{s(t)} Q_{s(t)} = R(t).
\end{equation}
From here we find the solution at time $t$:

\begin{equation}
Q_{s(t)} = \Theta^{\dagger}_{s(t)}R(t).
\end{equation}
The weights are updated as:
\begin{equation}
W_{s(t)}(t+1) = W_{s(t)}(t) + Q_{s(t)}.
\end{equation}
Here, $W_{s(t)}$ is the block of $W$ containing only the rows from the random list $s(t)$. Also, the new residual 
is updated as following:
\begin{equation}
R(t+1) = R(t) - \Theta_{s(t)} Q_{s(t)}.
\end{equation}
Due to the orthogonality of these quantities we have:
\begin{equation}
\Vert R(t+1) \Vert^2 = \Vert R(t) \Vert^2 - \Vert \Theta_{s(t)} Q_{s(t)} \Vert^2 \leq \Vert R(t) \Vert^2,
\end{equation}
and the algorithm is guaranteed to converge. 
Once the matrix $W$ is calculated, one can use the multi-class classifier defined by the equations (14)-(15), 
in order to classify any new $x \in \mathbb{R}^M$ data sample.

\subsection{Randomized Kaczmarz-MP method}

Interestingly, one can easily combine the Kaczmarz and the MP methods into a hybrid Kaczmarz-MP kernel method. 
In both methods we have to select a random list $s(t)$ of $J$ indexes, therefore we can use the same list to perform a Kaczmarz step 
followed by an MP step (or vice versa). The cost of this approach is similar to the randomized Nystr\"{o}m method, since it 
requires two matrix pseudo-inverses per iteration step. 

\section{Numerical results}

In this section we provide several new numerical results obtained for the MNIST\cite{key-13} and the CIFAR-10\cite{key-14} data sets, 
that illustrate the practical applicability and the performance of the proposed methods.

\subsection{MNIST data set}

The MNIST data set of hand-written digits is a a widely-used benchmark for testing multi-class classification algorithms. 
The data consists of $N=60,000$ training samples and 10,000 test samples, corresponding to $K=10$ different classes. 
All the images $x_i$, $i=1,\dots,N$, are of size $28\times 28 = 784$ pixels, with $256$ gray levels, and the intra-class variability consists in local translations 
and deformations of the hand written digits. Some typical examples of images from the MNIST data set are shown in Figure 1. 

While the MNIST data set is not very large, it still imposes a challenge for the standard kernel classification approach based on LS-SVM, because 
the kernel matrix has $36\cdot 10^8$ elements, which in double precision (64bit floating point numbers) 
requires about 28.8 GB. 
In order to simulate even more drastic conditions we used a PC with only 8 GB of RAM, 
and 4 CPU cores. Also, all the simulation programs were written using the Julia language.\cite{key-15}
Obviously, with such a limited machine, the attempt to solve the problem directly (for example using the Gaussian elimination or 
the Cholesky decomposition) is not feasible. However, we can easily use the proposed randomized approximation methods, 
and in order to do so we perform all the computations in single precision (32 bit floating point numbers), 
which provides a double storage capability, comparing to the case of double precision. 
Thus, we trade a potential precision loss for extra memory storage space, in order to adapt to the data size.  

In our numerical experiments we have used only the raw data, without any augmentation or distortion. 
It is well known that better results can be obtained by using more complex unsupervised learning of image features, 
data augmentation and image distortion methods at the pre-processing level. 
However, our goal here is only to evaluate the performance of the discussed methods, and therefore we prefer 
to limit our investigation to the raw data.

To our knowledge, the best reported results in the literature for the kernel SVM classification of the MNIST raw data have a classification error 
of $1.4\%$,\cite{key-16} and respectively $1.1\%$,\cite{key-3} and they have been obtained by combining ten kernel SVM classifiers. 
We will use these values for comparison with our results.

\begin{figure}[!t]
\centering \includegraphics[width=12cm]{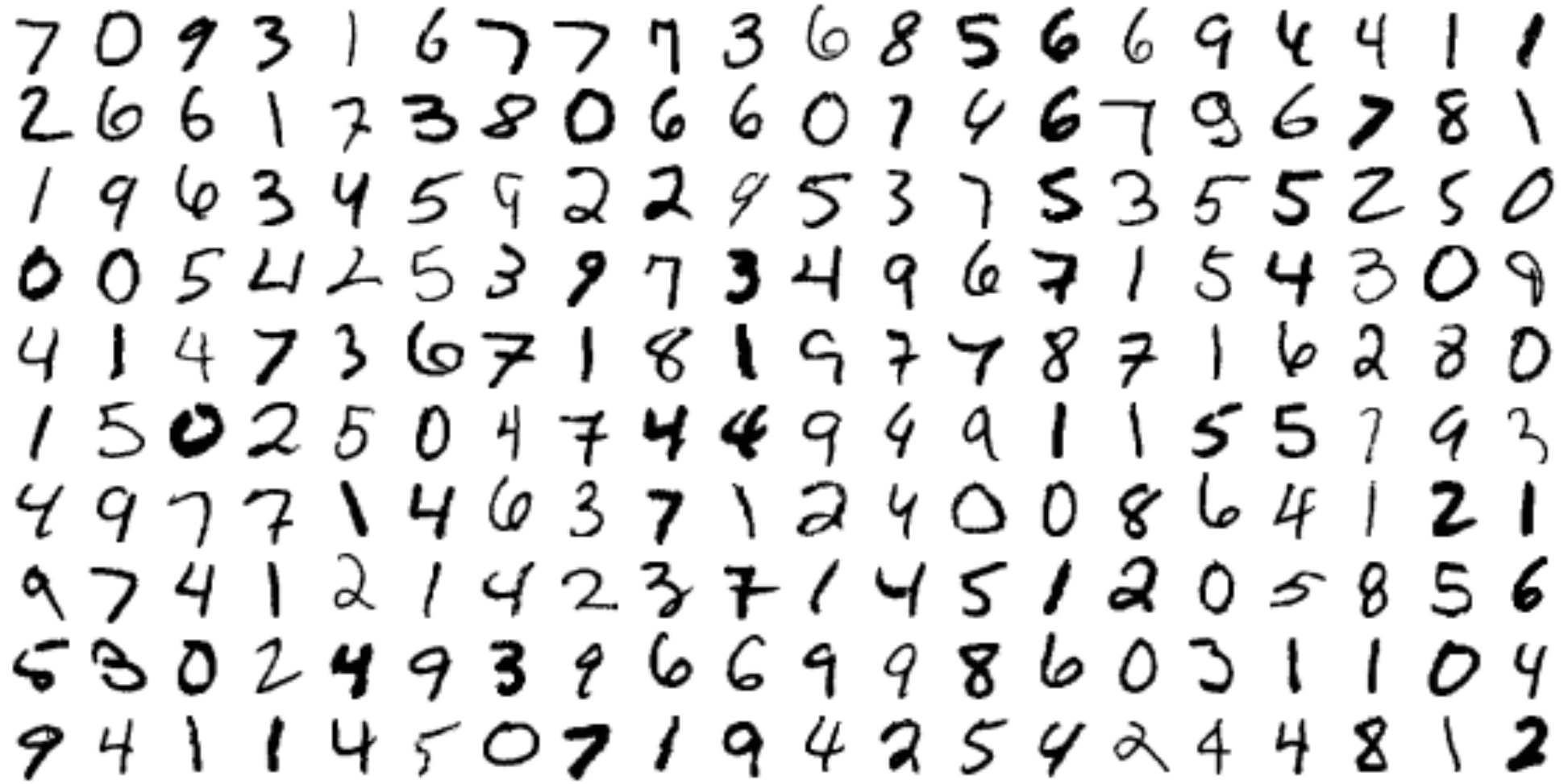}\caption{The first 200 hand-written digits from the MNIST data set.}
\end{figure}

In our approach we used a simple pre-processing method, consisting in a two step normalization of each image, as following:
\begin{align}
x_i &\leftarrow x_i - \langle x_i \rangle, \\
x_i &\leftarrow x_i/\Vert x_i \Vert, 
\end{align}
where $\langle \cdot \rangle$ denotes the average. Also, the images are "vectorized" by concatenating the columns, such that: $x_i \in \mathbb{R}^{784}$.
Therefore, after pre-processing all the images are unit norm vectors: $\Vert x_i \Vert = 1$, $i=1,\dots,N$. 
This normalization is useful because it makes all the inner products equal to the cosine of the angle between the vectors, which is a good similarity measure: 
\begin{equation}
\langle x_i, x_j \rangle = \cos (x_i,x_j) \in [-1,1], \quad \forall i,j=1,\dots,N.
\end{equation}

We have experimented with several kernel types (Gaussian, polynomial), and the best results have been obtained with 
a polynomial kernel of degree 4:
\begin{equation}
\Omega(x,x') = \langle x, x' \rangle^4.
\end{equation}
Therefore all the results reported here are for this particular kernel function. Also, the regularization parameter was always set to $\gamma = 10^{4}$, 
and the classification error $\eta$ was simply measured as the percentage of the test images which have been incorrectly classified.

The full kernel matrix: 
\begin{equation}
\Omega = (X_{train}X^T_{train})^{\{4\}},
\end{equation}
would still require 14.4 GB, which is not feasible with the imposed constraints, and therefore we must calculate the kernel elements on the fly at each step. 
Here, the exponent $\{4\}$ means that the power is calculated element-wise. Also we assume that the vectorized images correspond to the rows of the matrix $X_{train}$. 

The memory left is still enough to hold the matrix:
\begin{equation}
\Psi = (X_{test}X^T_{train})^{\{4\}},
\end{equation}
which is required for the classification of the testing data. This matrix is not really necessary to store in memory since the classification can be done separately for 
each test image, once the weights and the biases have been computed. This matrix requires about 2.4 GB and it is convenient to store it just 
to be able to perform a fast measurement of the classification error after each iteration. 

\begin{figure}[!t]
\centering \includegraphics[width=15cm]{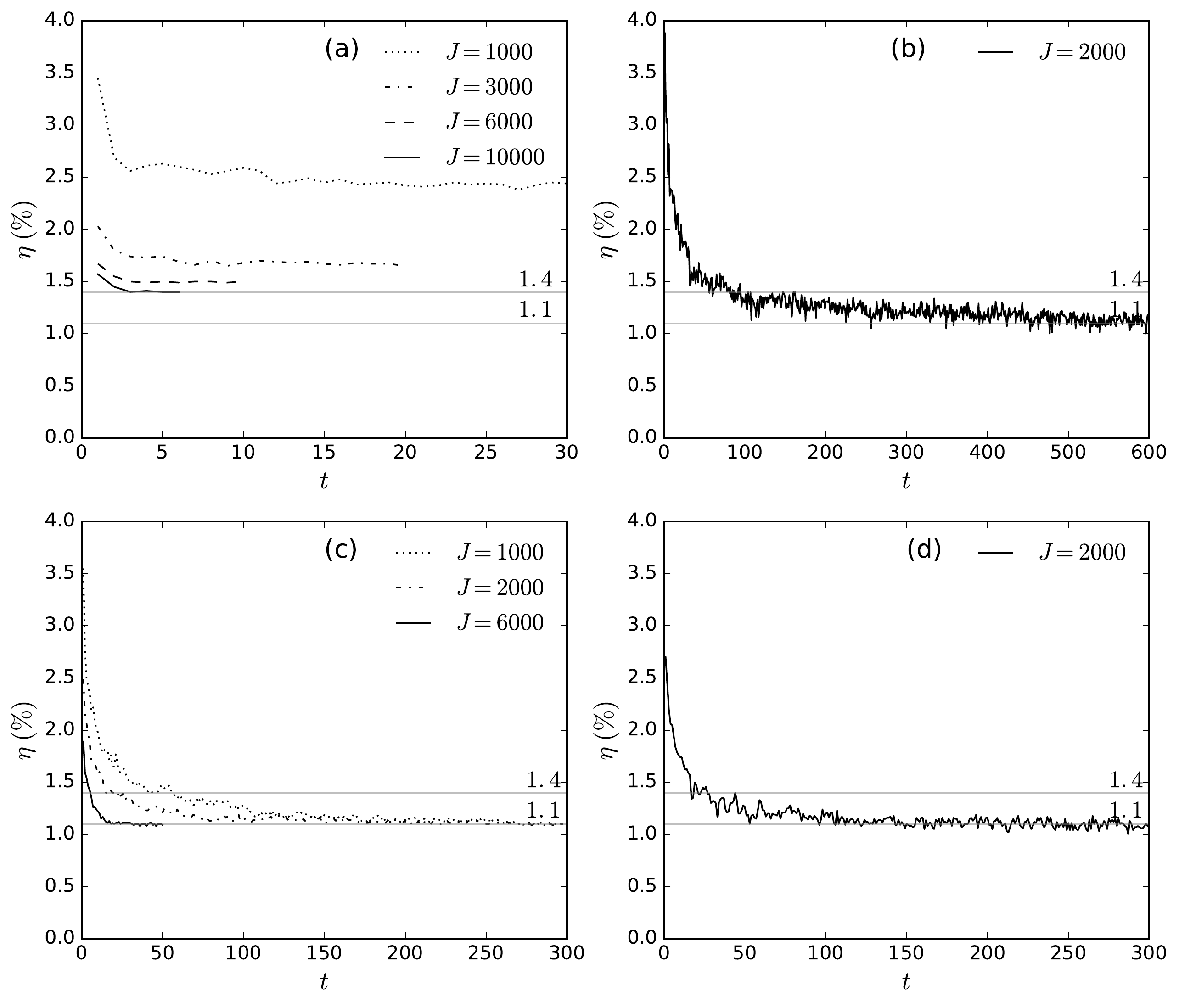}\caption{
The time evolution of the classification error $\eta$ for the iterative randomized kernel methods (MNIST data set): 
(a) Nystr\"{o}m method; (b) Kaczmarz method; (c) MP method; (d) Kaczmarz-MP method. Here $t$ is the 
iteration time step and $J$ is the size of the random block submatrices. 
}
\end{figure}

\begin{figure}[!t]
\centering \includegraphics[width=10cm]{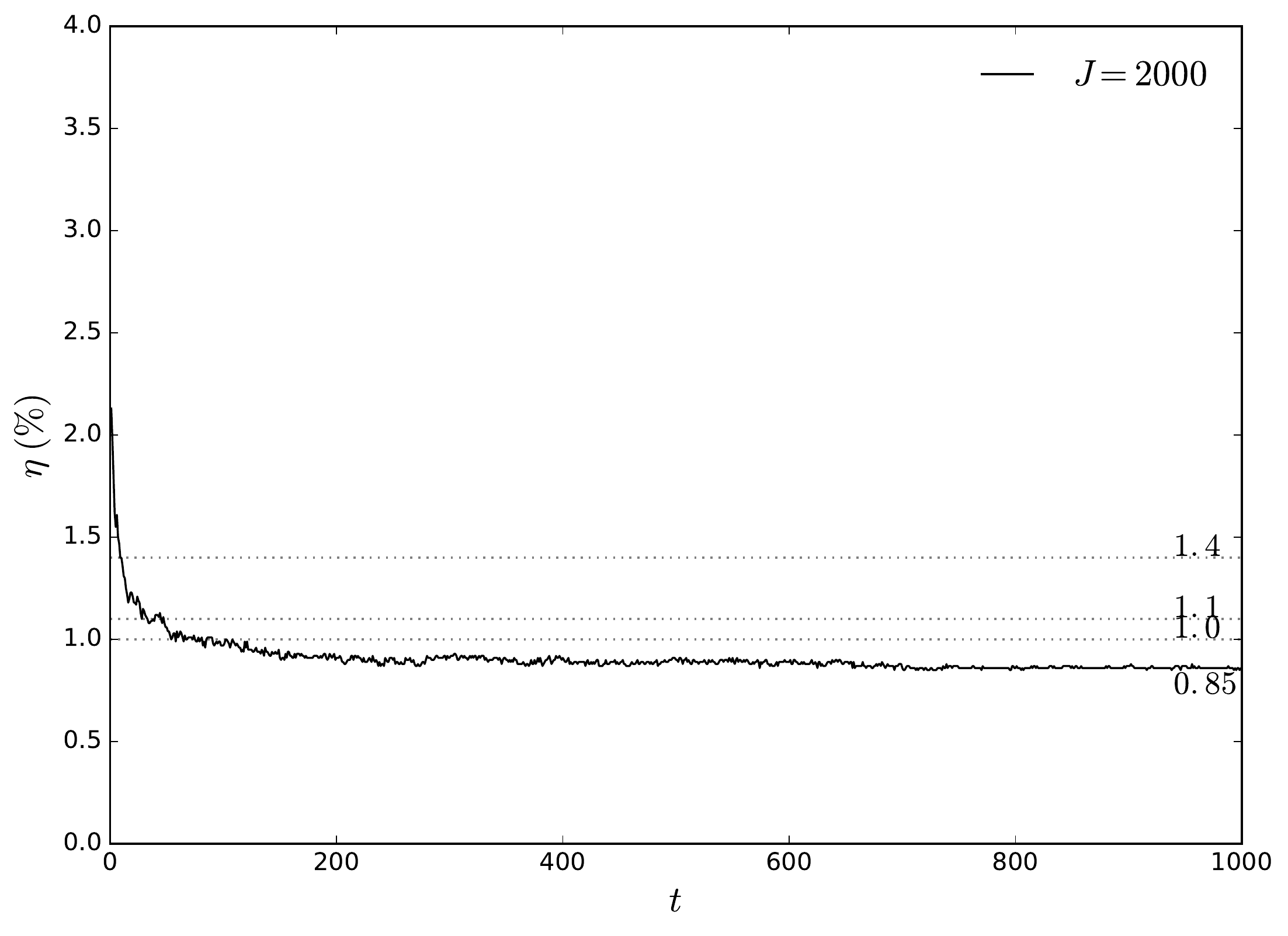}\caption{
The time evolution of the classification error $\eta$ for the randomized MP method 
with the Fourier pre-processing (MNIST data set). Here $t$ is the 
iteration time step and $J$ is the size of the random block submatrices.}
\end{figure}

In Figure 2(a) we give the results obtained for the randomized Nystr\"{o}m method. 
Here we have represented the classification error as a function of the random block size $J$ and the 
number of aggregated classifiers $t$. One can see that the method converges very fast, and only few classifiers are needed to reach a plateau for $\eta$. 
Unfortunately, the error $\eta$ is dependent on $J$, and this result suggests that the method performs better for larger values of $J$. 
This is the fastest method considered here, since it requires the aggregation of only a few classifiers, such that for $J=10,000$ the 
total number of necessary classifiers is only $T=6$. 

The randomized Kaczmarz method shows a different behavior, Figure 2(b). 
In this case we ran the algorithm for $T=600$ iterations using random data blocks of size $J=2000$. 
This method shows a much slower convergence, and the whole computation process needed almost 5 hours to complete 
($29$ seconds per iteration time step), including the time for the error evaluation at each step. 
After the first iteration step the classification error drops abruptly below $4\%$, 
then the classification error decreases slowly, and fluctuates closer to  
$\eta \simeq 1.1 \pm 0.1 \%$, which is in very good agreement with the previously reported values.

Better results for the Kaczmarz method can be obtained by simply averaging the weights and biases of several separately (parallel) trained classifiers, 
with different random seeds for the random number generator, and/or different block sizes $J$. 
This form of aggregation is possible because the kernel matrix is the same for all classifiers, and the 
systems are linear in the dual space. Assuming that we have $S$ classifiers, $s=1,\dots,S$, each of them having the output:
\begin{equation}
h_j^{(s)}(x) = \sum_{n=1}^N \Omega(x, x_n) a^{(s)}_{nj} + b^{(s)}_j, \quad j=1,\dots,K, \; s=1,\dots,S,
\end{equation}
then the output of the averaged classifiers is:
\begin{equation}
\hat{h}_j(x) = \sum_{n=1}^N \Omega(x, x_n) S^{-1}\sum_{s=1}^S a^{(s)}_{nj} + S^{-1}\sum_{s=1}^S b^{(s)}_j, \quad j=1,\dots,K.
\end{equation}
Therefore, in the end one can store only the average values of the weights and the biases. The advantage of averaging is the decrease of the 
amplitude of the fluctuations in the classification error, which gives better and more confident results, 
and also increases the generalization capabilities. 

In Figure 2(c) we give the results for the randomized MP method. This method shows similar results to the randomized Kaczmarz method, 
but it's convergence is faster and the fluctuations have a lower amplitude. Also, the method is about twice faster, 
with $14$ seconds per iteration time step, for a random block of size of $J=2000$. Again, the obtained result 
$\eta \simeq 1.1 \pm 0.02 \%$ is in very good agreement with the previously reported results. 

The results for the hybrid Kaczmarz-MP method are given in Figure 2(d), and not surprisingly it shows inbetween convergence speed, 
and a classification error of $\eta \simeq 1.1 \pm 0.05 \%$, which confirms the previous results.

In our last experiment we have decided to modify the data pre-processing, in order to see if better results can be obtained.
We only did a simple modification by concatenating the images with the square root of the absolute value of their fast Fourier transform ($\text{FFT}$). 
Since the FFT is symmetrical, only the first half of the values were used, each image becoming a vector of 1176 elements.

The new image pre-processing consists in the following steps:
\begin{align}
\xi_i &\leftarrow x_i, \\
\xi_i &\leftarrow \xi_i - \langle \xi_i \rangle \\
\varphi_i &\leftarrow \vert \text{FFT}(\xi_i) \vert^{1/2}, \\
\varphi_i &\leftarrow \varphi_i - \langle \varphi_i \rangle \\
\xi_i &\leftarrow \xi_i/\Vert \xi_i \Vert, \\
\varphi_i &\leftarrow \varphi_i/\Vert \varphi_i \Vert, \\
x_i &\leftarrow \frac{1}{\sqrt{2}} [\xi_i, \varphi_i]^T, \quad i=1,\dots,N.
\end{align}
With this new pre-processing we used the randomized MP method and the results are shown in Figure 3. 
The classification error drops to $\eta \simeq 0.85 \pm 0.01 \%$, which means an improvement of $0.25 \%$ over the previously reported results. 
The 85 images incorrectly recognized are shown in Figure 4.

\begin{figure}[!h]
\centering \includegraphics[width=10cm]{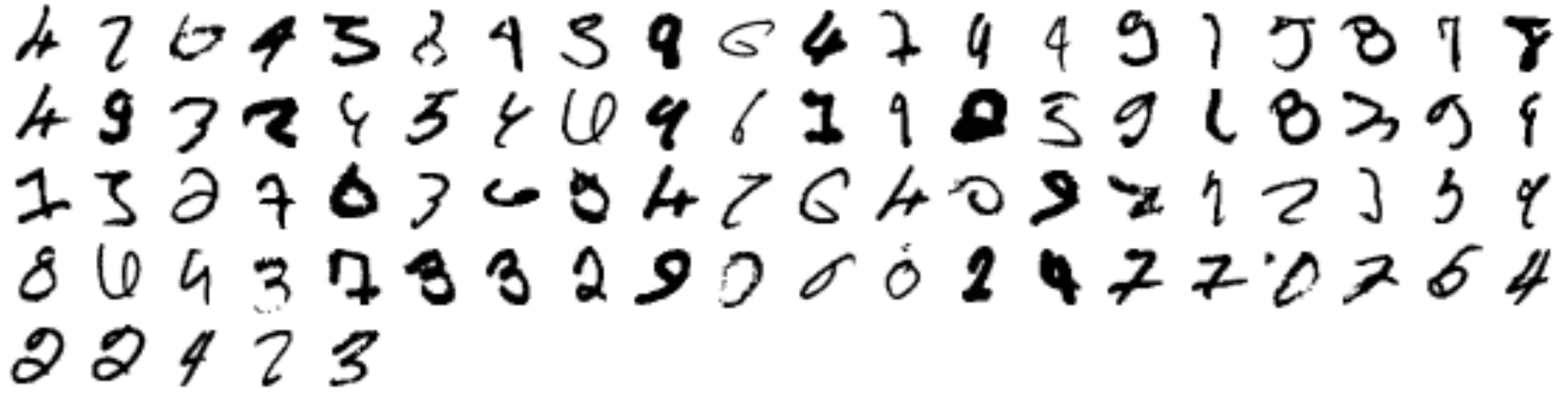}\caption{
The 85 images incorrectly recognized from the MNIST data set.}
\end{figure}

\subsection{CIFAR-10 data set}

The CIFAR-10 dataset consists of 60,000 images. 
Each image has $32 \times 32 = 1024$  pixels and 3 colour (RGB) channels with 256 levels. 
The data set contains 10 classes of equal size (6,000), 
corresponding to objects (airplane, automobile, ship, truck) 
and animals (bird, cat, deer, dog, frog, horse). 
The training set contains 50,000 images, while the test set contains the rest of 10,000 images.

\begin{figure}[!t]
\centering \includegraphics[width=11cm]{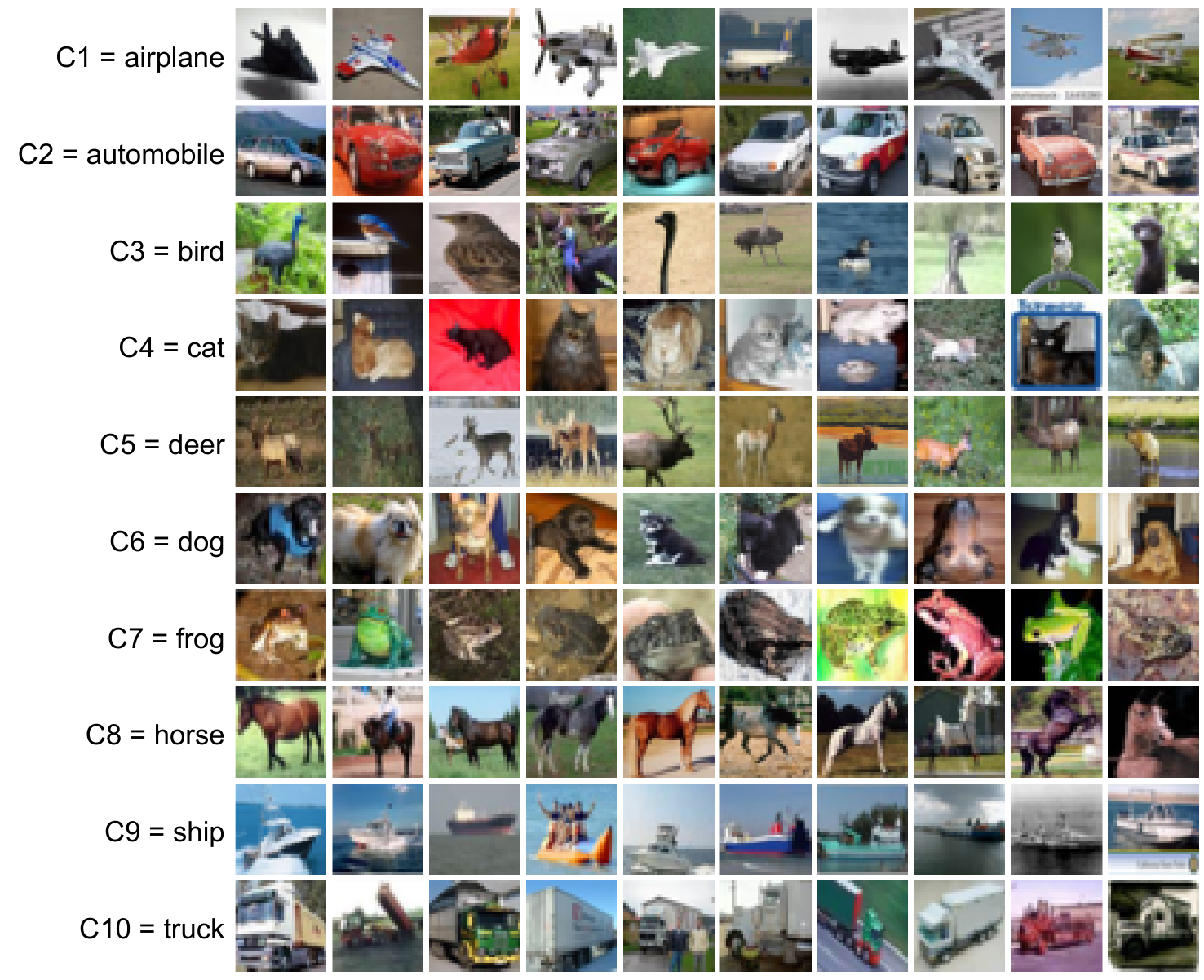}\caption{
The first 10 images for each class from the CIFAR-10 data set.}
\end{figure}

In Figure 5 we show the first 10 images from each class. 
Obviously this data set is much more challenging than the MNIST data set, showing a very high  
variation in each class. 
Again, we only use the raw data without any augmentation and distortion, 
and without employing any other technique for advanced feature extraction and learning. 
In Figure 6 we give the results of the numerical experiments for the randomized MP method, 
with the random data block size $J=2500$. 

In the first experiment ($E1$) we used the simple two-step normalization described in (61)-(62). 
Thus, after the normalization steps each image becomes a unit vector with $3072$ pixels, obtained by concatenating the columns 
from the three colour channels. 
We ran the randomized MP algorithm for $T=100$ steps, such that 
the classification error settled at $\eta = 43.93 \pm 0.03 \%$. This is 
a good improvement over the random guessing "method", which has an error $\eta = 90 \%$. 

In the second experiment ($E2$), before the two-step normalization 
we have applied a Gaussian filter on each channel. The filter is centred in the middle 
of the image and it is given by:
\begin{equation}
G_{ij} = \exp \left( -c[(i-L/2)^2 + (j-L/2)^2] \right) , \; i,j=1,\dots,L,
\end{equation}
where $L = 32$ is the size of the image. The filter is applied element-wise to the pixels and it is used 
to enhance the center of the image, where supposedly the important information resides, 
and to attenuate the information at the periphery of the image. The best results 
have been obtained for a constant $c=4L^{-2}$, and the classification error dropped to $\eta = 41.21 \pm 0.03 \%$. 

In the third experiment ($E3$) we have used the Gaussian filtering (76) and the FFT normalization described by the equations (69)-(75). 
This pre-processing method improved the 
results again and we have obtained an error of only $\eta = 33.71 \pm 0.01 \%$.
The resulted confusion matrix is given in Table 1.  
Not surprisingly, one can see that main "culprits" are the cats and dogs. For $20.9 \%$ the cats are mistaken for dogs, while for $18.3 \%$ 
the dogs are mistaken for cats. 

\begin{figure}[!t]
\centering \includegraphics[width=9cm]{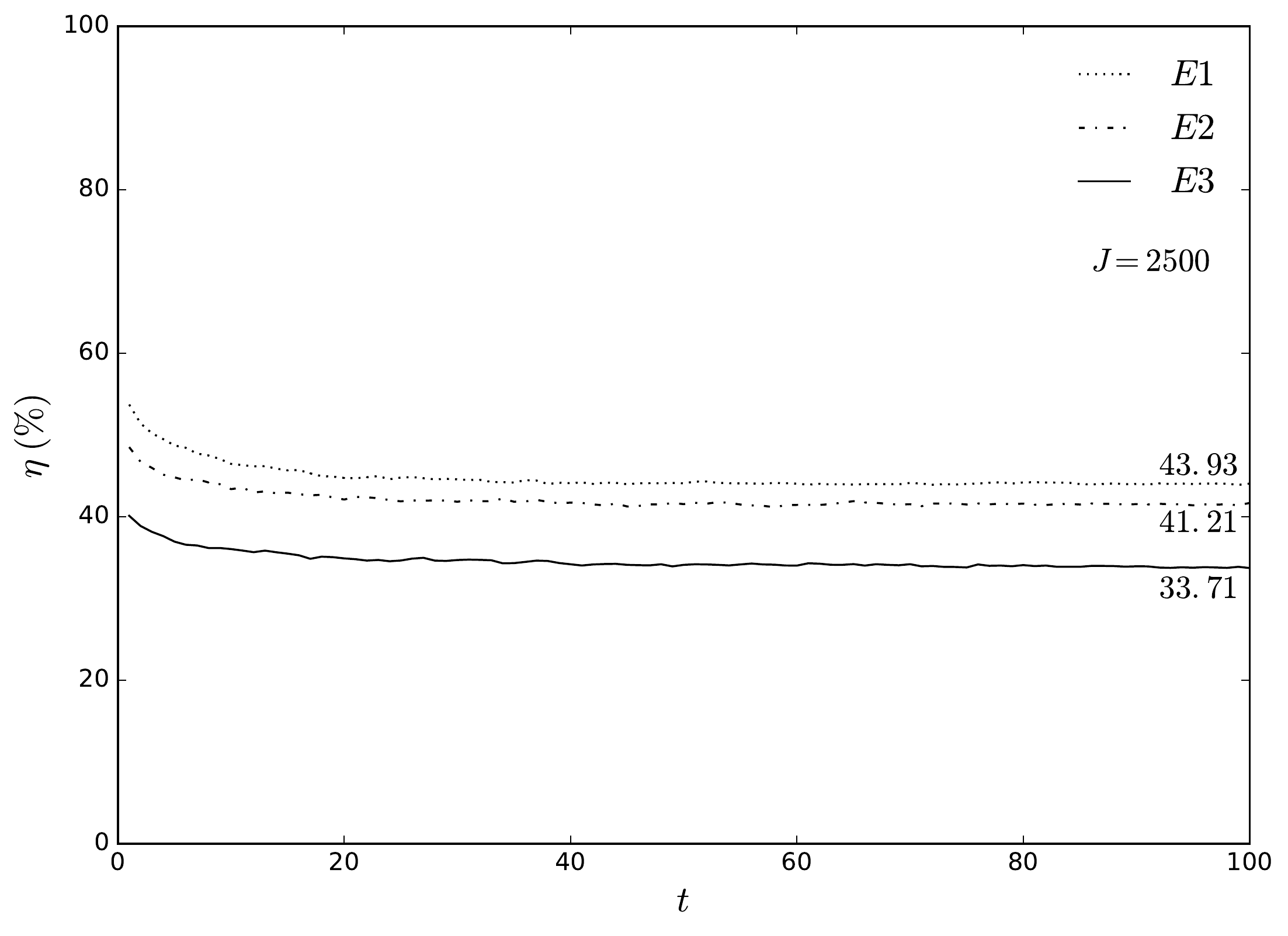}\caption{
The time evolution of the classification error $\eta$ for the randomized MP method (CIFAR-10 data set). 
Here $t$ is the iteration time step and $J$ is the size of the random block submatrices. The following 
pre-processing steps have been used:  
$E1$ - two-step data normalization (61)-(62); $E2$ - Gaussian filtering (76) and two-step data normalization (61)-(62); 
$E3$ - Gaussian filtering (76) and FFT normalization (69)-(75).
}
\end{figure}

\begin{table}

\caption{The confusion matrix ($\%$) for the CIFAR-10 data set and the classification error $\eta = 33.71 \%$.}
\bigskip
\centering
{
\begin{tabular}{l*{10}{c}r}
Class & C1 & C2 & C3 & C4 & C5 & C6 & C7 & C8 & C9 & C10 \\
\hline
C1 & 71.0 & 2.3 & 4.5 & 1.9 & 2.2 & 2.0 & 1.4 & 1.4 & 9.6 & 3.7 \\
C2 & 3.1 & 75.7 & 1.1 & 1.2 & 1.1 & 0.6 & 1.0 & 1.0 & 4.4 & 10.8 \\
C3 & 6.8 & 1.1 & 54.5 & 6.4 & 8.6 & 7.0 & 8.5 & 4.1 & 1.8 & 1.2 \\
C4 & 2.3 & 1.8 & 6.6 & 44.5 & 6.2 & 20.9 & 8.5 & 4.0 & 2.3 & 2.9 \\
C5 & 3.1 & 1.0 & 7.5 & 7.3 & 56.1 & 5.8 & 7.2 & 7.9 & 2.4 & 1.7 \\
C6 & 1.1 & 0.7 & 5.7 & 18.3 & 5.8 & 58.3 & 2.5 & 4.6 & 1.2 & 1.8 \\
C7 & 1.1 & 0.7 & 4.8 & 4.6 & 4.9 & 3.1 & 78.2 & 1.1 & 0.6 & 0.9 \\
C8 & 2.1 & 1.5 & 3.3 & 5.3 & 5.8 & 7.0 & 1.8 & 69.3 & 1.0 & 2.9 \\
C9 & 6.8 & 5.3 & 1.1 & 0.9 & 1.0 & 1.0 & 0.9 & 0.5 & 78.2 & 4.3 \\
C10 & 2.6 & 8.8 & 1.2 & 2.6 & 0.5 & 0.7 & 0.2 & 0.8 & 5.5 & 77.1 \\
\end{tabular}
}
\end{table}

\section{Conclusion}

In this paper we have discussed several approximation methods for the kernel LS-SVM multi-class classification problem. 
These methods use randomized block kernel matrices,  
and their main advantages are the low memory requirements and the relatively simple iterative implementation, 
significantly reducing the complexity of the problem. Another advantage is that these methods do not require 
complex parameter tuning mechanisms. The only parameters needed are 
the regularization parameter and the size of the random block matrices. 
Despite of their simplicity, these methods provide very good accuracy and reliable scaling to 
relatively large multi-class classification problems. 
Also, we have reported new numerical results for the MNIST and CIFAR-10 data sets,  
and we have shown that better results can be obtained by using a simple 
Fourier pre-processing step of the raw data.
The results reported here for the MNIST raw data set are in very good agreement with the 
previously reported results using the kernel SVM approach, while the results for the CIFAR-10 data are  
surprisingly good for the small amount of pre-processing used.


\begin{thebibliography}{00}

\bibitem{key-1} T. Hofmann, B. Sch\"{o}lkopf, A. J. Smola, 
				\textit{Kernel methods in machine learning, }
				The Annals of Statistics 36(3), 1171 (2008).
\bibitem{key-2} J. Mercer, 
				\textit{Functions of positive and negative type and their connection with the theory of integral equations,}
				Philos. Trans. R. Soc. Lond. Ser. A Math. Phys. Eng. Sci. A 209, 415 (1909). 
\bibitem{key-3} C. Cortes, V. Vapnik,
				\textit{Support Vector Networks}, 
				Machine Learning 20, 273 (1995).
\bibitem{key-4} J.A.K. Suykens, J. Vandewalle, 
				\textit{Least squares support vector machine classifiers}, 
				Neural Processing Letters 9(3), 293 (1999).
\bibitem{key-5}	E. J. Nystr\"{o}m, 
				\textit{\"{U}ber die praktische Aufl\"{o}sung von Integralgleichungen mit Anwendungen auf Randwertaufgaben",}
				Acta Mathematica 54(1), 185 (1930).
\bibitem{key-6} S. Kaczmarz, 
				\textit{Angen\"{a}herte Aufl\"{o}sung von Systemen linearer Gleichungen,} 
				Bulletin International de l'Acad\'{e}mie Polonaise des Sciences et des Lettres. 
				Classe des Sciences Math\'{e}matiques et Naturelles. S\'{e}rie A, Sciences Math\'{e}matiques, 35, 355 (1937).
\bibitem{key-7} S. G. Mallat, Z. Zhang, 
				\textit{Matching Pursuits with Time-Frequency Dictionaries,} 
				IEEE Transactions on Signal Processing 41(12), 3397 (1993).
\bibitem{key-8} C.K.I Williams, M. Seeger, 
				\textit{Using the Nystr\"{o}m Method to Speed Up Kernel Machines,} 
				Proceedings Neural Information Processing Systems 13, 682 (2001).
\bibitem{key-9} J. Sherman, W.J. Morrison,  
				\textit{Adjustment of an Inverse Matrix Corresponding to a Change in One Element of a Given Matrix, } 
				Annals of Mathematical Statistics 21, 124 (1950).
\bibitem{key-10} M. Girolami, 
				\textit{Orthogonal series density estimation and the kernel eigenvalue problem, }
				Neural Computation 14, 669 (2002).
\bibitem{key-11} T. Strohmer, R. Vershynin, 
				\textit{A randomized Kaczmarz algorithm for linear systems with exponential convergence}, 
				Journal of Fourier Analysis and Applications 15, 262 (2009). 
\bibitem{key-12} P. Vincent, Y. Bengio, 
				\textit{Kernel Matching Pursuit,} 
				Machine Learning 48, 169 (2002).
\bibitem{key-13}	Y. Lecun, L. Bottou, Y. Bengio, P. Haffner, 
				\textit{Gradient-based learning applied to document recognition,} 
				Proceedings of the IEEE 86(11), 2278 (1998).
\bibitem{key-14}	A. Krizhevsky, G. Hinton. 
				\textit{Learning multiple layers of features from tiny images,} Computer Science Department, University of Toronto, Tech. Rep., 2009.			
\bibitem{key-15} J. Bezanson, S. Karpinski, V.B. Shah, A. Edelman, 
				\textit{Julia: A Fast Dynamic Language for Technical Computing}, 
				arXiv:1209.5145 (2012).
\bibitem{key-16} C.J.C. Burges, B. Sch\"{o}lkopf, 
				\textit{Improving the Accuracy and Speed of Support Vector Machines}, 
				Advances in Neural Information Processing Systems 9, 375 (1997).
\end{thebibliography}
\end{document}